\documentclass[conference]{IEEEtran}
\IEEEoverridecommandlockouts

\usepackage{cite}
\usepackage{amsmath,amssymb,amsfonts,bm,mathtools}
\usepackage{algorithmic}
\usepackage{booktabs}
\usepackage{multirow}
\usepackage{makecell}
\usepackage{pifont}
\usepackage{graphicx}
\usepackage{textcomp}
\usepackage{xcolor}
\usepackage{url}
\usepackage{placeins}
\usepackage{float}
\usepackage{needspace}
\usepackage{balance}

\newcommand{\BibTeX}{\mathrm{B\kern-.05em{\sc i\kern-.025em b}\kern-.08em
T\kern-.1667em\lower.7ex\hbox{E}\kern-.125emX}}

\newcommand{\simf}{\operatorname{sim}}

\newcommand{\best}[1]{\textbf{#1}}
\newcommand{\second}[1]{\underline{#1}}
\newcommand{\reimpl}{\textsuperscript{\dag}}

\begin{document}
\flushbottom

\title{Reliability-Role Collaborative Learning for Semi-Supervised Cardiac Image Segmentation%

\thanks{This work was supported by the National Key Research and Development Program of China under Grant 2025YFC2708700, in part by the Brain-Gain Plan of New Chongqing Foundation under Grant CSTB2024YCJH-KYXM0108, in part by the Natural Science Foundation of Chongqing under Grant CSTB2025NSCQ-GPX1037, in part by the Natural Science Foundation Innovation and Development Joint Fund Project of Chongqing under Grant CSTB2023NSCQ-LZX0148, in part by the Science and Technology Research Program of Chongqing Municipal Education Commission under Grants KJQN202500532 and KJQN202500511, and in part by Wuwei Tumour Hospital of Gansu Province under Grant WW25Z01SF075.}%
\thanks{Zhongrui Xie and Wenfeng Zhang contributed equally to this work.}%
\thanks{Corresponding authors: Yi Zhang (zhangyii@cqnu.edu.cn); Tao Jia (tjia@cqnu.edu.cn).}%

}

\author{
\IEEEauthorblockN{
Zhongrui Xie\textsuperscript{1,$\ast$},
Wenfeng Zhang\textsuperscript{1,$\ast$},
Wei Hu\textsuperscript{1},
Ruitao Xie\textsuperscript{2},
Wenjie Wang\textsuperscript{3},
Zhouyun Xie\textsuperscript{1},
Yi Zhang\textsuperscript{1,$\dagger$},
Tao Jia\textsuperscript{1,$\dagger$}
}
\IEEEauthorblockA{
\textsuperscript{1}College of Computer and Information Science,\\
Research Center of Biomimetic Intelligence and Computational Biomedicine,\\
Chongqing Normal University, Chongqing, China\\
\textsuperscript{2}Shenzhen University of Advanced Technology, Shenzhen, China\\
\textsuperscript{3}Wuwei Tumour Hospital of Gansu Province, Wuwei, China
}
}

\maketitle

\begin{abstract}
Semi-supervised cardiac image segmentation aims to exploit abundant unlabeled images together with limited annotations, but the reliability of pseudo-label supervision is highly uneven across spatial regions.
In a peer dual-network collaborative setting, interior regions are usually more stable, whereas boundary regions, low-confidence areas, and confidence-discrepant regions between the two peer networks are more likely to introduce noisy supervision.
However, despite this spatial reliability heterogeneity, existing methods typically impose uniform supervision on all unlabeled pixels, overlooking reliability differences across regions and consequently limiting discriminative learning in challenging areas.
To address this issue, we propose R2CL (Reliability-Role Collaborative Learning), a reliability-role collaborative learning framework for semi-supervised cardiac image segmentation.
R2CL first performs reliability-role modeling based on confidence relationships between two networks, assigning different training roles to unlabeled regions: mutually uncertain regions are used for entropy regularization, confidence-discrepant regions for directional knowledge transfer, and mutually confident regions as reliable supervision anchors.
Based on this formulation, Reliability-Role Mixed Cross Learning (RR-MCL) operationalizes these learning roles through region-mixed training, directional cross-network supervision, and entropy regularization over mutually uncertain regions.
Subsequently, Reliability-Transition Prototype Contrastive Learning (RT-PCL) constructs query features from foreground-adjacent reliability-transition regions and learns network-specific class prototypes from mutually confident regions, thereby enhancing feature discriminability around ambiguous boundaries and difficult regions.
Meanwhile, High-confidence Context Refinement (HCR) further exploits reliable interior pixels and trusted dense blocks within mutually confident regions to complement stable contextual supervision.
R2CL achieves stable improvements in region-overlap and boundary-related metrics across multiple datasets, with more evident advantages on the foreground-sparse MVAA-CT dataset.
\end{abstract}

\begin{IEEEkeywords}
cardiac segmentation, semi-supervised learning, contrastive learning, mitral valve segmentation
\end{IEEEkeywords}

\section{Introduction}

Medical image segmentation supports computer-aided diagnosis, quantitative image analysis, and clinical planning.
CNN-based models such as U-Net~\cite{ronneberger2015unet} and nnU-Net~\cite{isensee2021nnunet}, together with Transformer architectures including TransUNet~\cite{chen2024transunet} and Swin UNETR~\cite{hatamizadeh2022swinunetr}, have advanced anatomical segmentation.
However, their performance relies heavily on expert pixel-wise annotations, which are particularly costly for volumetric images.
Semi-supervised learning alleviates this burden by exploiting abundant unlabeled data alongside limited annotations.

Existing semi-supervised learning approaches include pseudo-label learning~\cite{lee2013pseudo}, Mean Teacher~\cite{tarvainen2017mean}, and cross pseudo supervision~\cite{chen2021cps}, often combined with consistency regularization~\cite{miyato2019vat} and data mixing~\cite{berthelot2019mixmatch}.
In medical image segmentation, representative methods such as UAMT~\cite{yu2019uamt}, MC-Net~\cite{wu2022mcnet}, BCP~\cite{bai2023bcp}, and ALHVR~\cite{lei2025alhvr} further incorporate uncertainty estimation, perturbation consistency, copy-paste mixing, and high-value region utilization to alleviate annotation scarcity.
Nevertheless, these methods to some extent implicitly assume homogeneous supervision across unlabeled pixels, thereby overlooking differences in reliability and learning roles among spatial regions.

However, the reliability of unlabeled predictions is inherently heterogeneous across spatial regions.
Interior regions of anatomical structures generally exhibit stable predictions, whereas boundary regions, low-confidence areas, and disagreement regions between dual networks are more susceptible to class ambiguity and noisy pseudo-label supervision.
This issue is particularly pronounced in cardiac MRI and mitral valve CT segmentation, where local false positives, structural discontinuities, and boundary deviations may compromise downstream clinical analyses, including ventricular volume estimation~\cite{luo2018multiview,luo2020dcn}, biventricular quantification~\cite{luo2020ccn}, valve geometry assessment, and shape continuity analysis.
More importantly, existing semi-supervised segmentation methods often implicitly assume that all unlabeled pixels contribute equally to model optimization and therefore impose uniform consistency regularization or pseudo-label supervision across all regions.
Such a homogeneous supervision assumption overlooks the heterogeneous reliability characteristics of different spatial regions.
In contrast, different unlabeled regions naturally assume distinct learning roles: some are suitable for uncertainty suppression, some facilitate directional knowledge transfer, while others provide reliable supervisory signals.
Therefore, unlabeled learning should not merely be formulated as a uniform consistency regularization problem, but rather as a reliability-role learning problem.

Peer-network confidence relationships provide a basis for identifying different learning roles.
Inspired by complementary supervision in co-training~\cite{blum1998cotrain}, mutually uncertain regions can be used for entropy regularization, confidence-discrepant regions for directional knowledge transfer, and mutually confident regions as candidates for reliable supervision.
This role-based formulation allows different unlabeled regions to provide distinct training signals rather than relying solely on a single uniform supervisory objective.

Motivated by the above observations, we propose R2CL (Reliability-Role Collaborative Learning), a reliability-role collaborative learning framework for semi-supervised cardiac image segmentation.
R2CL first performs reliability-role modeling based on dual-network confidence relationships, assigning distinct learning roles to unlabeled regions.
RR-MCL then operationalizes these roles through region-mixed training, directional cross-network supervision, and entropy regularization.
Moreover, RT-PCL focuses on foreground-adjacent reliability-transition regions, where query features are contrasted against network-specific class prototypes to enhance feature discrimination around ambiguous structures.
HCR further exploits reliable interior pixels and trusted dense blocks within mutually confident areas to provide stable contextual supervision.
Together, these components integrate complementary supervisory signals from different reliability roles into a unified collaborative framework, enabling role-specific exploitation of heterogeneous unlabeled supervision.

Our main contributions are as follows:
\begin{list}{\textbullet}{%
\setlength{\leftmargin}{1.2em}
\setlength{\labelwidth}{0.6em}
\setlength{\labelsep}{0.5em}
\setlength{\itemsep}{0.18em}
\setlength{\topsep}{0pt}
\setlength{\parsep}{0pt}
\setlength{\partopsep}{0pt}}
\item We propose R2CL, a reliability-role collaborative learning framework for semi-supervised cardiac image segmentation. R2CL introduces a reliability-role modeling strategy that reformulates unlabeled learning as a reliability-role learning problem based on dual-network confidence relationships. By assigning distinct learning roles to unlabeled regions, it avoids uniform supervision across all unlabeled pixels.
\item We develop a role-specific collaborative learning mechanism that exploits complementary supervisory signals from different reliability roles through uncertainty suppression, reliability-transition learning, and high-confidence context refinement. This mechanism is instantiated by three complementary components, namely RR-MCL, RT-PCL, and HCR, enabling role-specific exploitation of heterogeneous unlabeled supervision.
\item Experiments on the ACDC and MVAA-CT datasets demonstrate improvements in region-overlap and boundary-related metrics, with particularly evident advantages in foreground-sparse mitral valve CT segmentation.
\end{list}

\begin{figure*}[!t]
\centering
\includegraphics[width=0.98\textwidth]{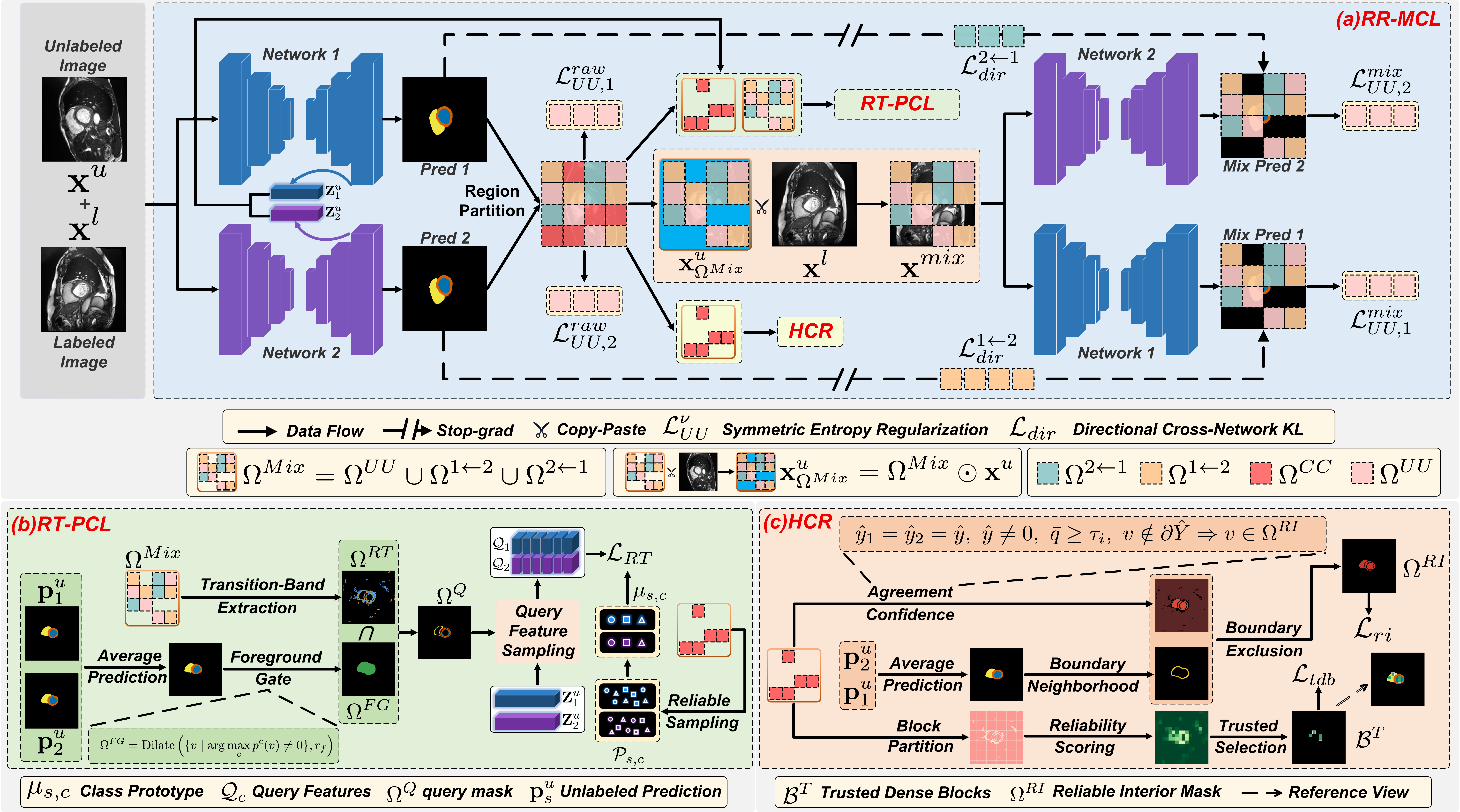}
\caption{Overall framework of R2CL. (a) RR-MCL constructs reliability-role mixed cross learning by converting unlabeled regions into region-mixed training, directional cross-network supervision, and entropy regularization signals. (b) RT-PCL performs foreground-adjacent reliability-transition query mining and network-specific class-prototype contrast. (c) HCR exploits reliable interior pixels and trusted dense blocks in mutually confident regions to enhance structural consistency.}
\label{fig:framework}
\end{figure*}

\section{Method}

\subsection{Problem Setting}

We consider semi-supervised segmentation with a small labeled set $\mathcal{D}_l=\{(x_i^l,y_i^l)\}_{i=1}^{N_l}$ and a larger unlabeled set $\mathcal{D}_u=\{x_j^u\}_{j=1}^{N_u}$, where $N_l\ll N_u$. Here, $x$ denotes an input image and $y$ denotes the corresponding segmentation annotation.

\subsection{Overview}

As shown in Fig.~\ref{fig:framework}, R2CL consists of three coupled modules: RR-MCL, RT-PCL, and HCR. RR-MCL assigns reliability roles to unlabeled regions and converts them into region-mixed training and directional cross-network supervision. RT-PCL constrains foreground-adjacent reliability-transition regions through network-specific class-prototype contrast. HCR further exploits mutually confident regions through reliable interior pixels and trusted dense blocks.
The two networks share the same encoder-decoder backbone but use different decoder upsampling operations, which encourages complementary local boundary recovery and confidence responses.

\subsection{Reliability-Role Mixed Cross Learning}

RR-MCL partitions unlabeled pixels according to peer-network confidence.
For a pixel $v$, let $p_s^c(v)$ denote the class-$c$ probability predicted by network $s$, with confidence $q_s(v)=\max_c p_s^c(v)$.
To avoid a fixed threshold, we use a mini-batch adaptive threshold based on the average confidence $\bar q(v)$, where $\mathcal{S}_{\rho}$ contains the lowest $\rho$ proportion of $\bar q(v)$ in the current mini-batch:

\begin{equation}
\bar{q}(v)=\frac{q_1(v)+q_2(v)}{2},\quad
\tau=\frac{1}{|\mathcal{S}_{\rho}|}\sum_{v\in\mathcal{S}_{\rho}}\bar{q}(v).
\label{eq:adaptive_tau}
\end{equation}
The threshold defines four reliability regions:

\begin{equation}
\begin{aligned}
\Omega^{UU}
&= \{v\mid q_1(v)<\tau,\ q_2(v)<\tau\},\\
\Omega^{1\leftarrow2}
&= \{v\mid q_1(v)<\tau,\ q_2(v)\ge\tau\},\\
\Omega^{2\leftarrow1}
&= \{v\mid q_1(v)\ge\tau,\ q_2(v)<\tau\},\\
\Omega^{CC}
&= \{v\mid q_1(v)\ge\tau,\ q_2(v)\ge\tau\}.
\end{aligned}
\label{eq:regions}
\end{equation}
Here, $\Omega^{UU}$, $\Omega^{1\leftarrow2}$, $\Omega^{2\leftarrow1}$, and $\Omega^{CC}$ denote mutually uncertain, network-2-guided, network-1-guided, and mutually confident regions, respectively.
The first three support region mixing, directional supervision, and transition mining, whereas $\Omega^{CC}$ provides reliable candidates for RT-PCL and HCR.
We construct the region-mixed input using the first three regions:

\begin{equation}
\begin{aligned}
&\Omega^{Mix}
= \Omega^{UU}\cup\Omega^{1\leftarrow2}\cup\Omega^{2\leftarrow1},\\
&\mathbf{x}^{u}_{\Omega^{Mix}}
= \Omega^{Mix}\odot\mathbf{x}^{u},\\
&\mathbf{x}^{mix} 
= \mathbf{x}^{u}_{\Omega^{Mix}}+(1-\Omega^{Mix})\odot\mathbf{x}^{l}.
\end{aligned}
\label{eq:mixinput}
\end{equation}
This mask-based composition preserves deterministic supervision in retained labeled regions and exposes selected unlabeled regions to role-specific constraints.
Labeled pixels outside $\Omega^{Mix}$ retain their ground-truth supervision:

\begin{equation}
\mathcal{L}_{lab}^{mix}
=
\sum_{s=1}^{2}
\mathbb{E}_{v\notin\Omega^{Mix}}
\left[
\ell_{seg}\!\left(\tilde{\mathbf p}_s(v),\mathbf y^l(v)\right)
\right],
\label{eq:mixlab}
\end{equation}
Here, $\tilde{\mathbf p}_s$ denotes the prediction of network $s$ on $\mathbf{x}^{mix}$; mixed unlabeled regions are constrained by the following role-specific objectives.

For the mutually uncertain region, neither network provides a sufficiently reliable supervisory signal.
We impose symmetric entropy regularization on this region from two views: the original unlabeled view and the region-mixed view.
The entropy constraint is defined as:

\begin{equation}
\begin{gathered}
\begin{aligned}
&\mathcal{L}_{UU}^{\nu}
= \mathcal{L}_{UU,1}^{\nu}+\mathcal{L}_{UU,2}^{\nu},
\quad \nu\in\{\mathrm{raw},\mathrm{mix}\},\\
&\mathcal{L}_{UU,s}^{\nu}
= \mathbb{E}_{v\in\Omega^{UU}}
\left[H\!\left(\mathbf{p}_{s}^{\nu}(v)\right)\right],
\quad s\in\{1,2\},
\end{aligned}\\
\mathbf{p}_{s}^{\mathrm{raw}}(v)=\mathbf{p}_s(v),
\qquad
\mathbf{p}_{s}^{\mathrm{mix}}(v)=\tilde{\mathbf{p}}_s(v).
\end{gathered}
\label{eq:uent}
\end{equation}
In confidence-discrepant regions, the higher-confidence network's raw prediction serves as a detached soft target for the lower-confidence network's mixed-view prediction:

\begin{equation}
\begin{aligned}
\mathcal{L}_{dir}^{1\leftarrow2}
&= \mathbb{E}_{v\in\Omega^{1\leftarrow2}}
D_{KL}\!\left(\operatorname{sg}[\mathbf{p}_2(v)]\Vert\tilde{\mathbf{p}}_1(v)\right),\\
\mathcal{L}_{dir}^{2\leftarrow1}
&= \mathbb{E}_{v\in\Omega^{2\leftarrow1}}
D_{KL}\!\left(\operatorname{sg}[\mathbf{p}_1(v)]\Vert\tilde{\mathbf{p}}_2(v)\right),\\
\mathcal{L}_{dir}
&= \mathcal{L}_{dir}^{1\leftarrow2}+\mathcal{L}_{dir}^{2\leftarrow1}.
\end{aligned}
\label{eq:dir}
\end{equation}
Here, $D_{KL}$ denotes KL divergence and $\operatorname{sg}[\cdot]$ stops gradients through the soft target.
This one-way gradient flow uses the high-confidence prediction as a detached soft target without updating it through the KL term.
The overall RR-MCL objective is:

\begin{equation}
\begin{aligned}
\mathcal{L}_{RR}
&= \lambda_{raw}(t)\mathcal{L}_{UU}^{raw}\\
&\quad+\lambda_{mix}(t)
\left(\mathcal{L}_{lab}^{mix}+\mathcal{L}_{dir}
+\lambda_{ll}\mathcal{L}_{UU}^{mix}\right).
\end{aligned}
\label{eq:rrloss}
\end{equation}
Here, $\mathcal{L}_{lab}^{mix}$ supervises retained labeled regions, $\mathcal{L}_{dir}$ performs directional soft-target transfer, and $\mathcal{L}_{UU}^{raw}$ and $\mathcal{L}_{UU}^{mix}$ are raw-view and mixed-view entropy terms.

\subsection{Reliability-Transition Prototype Contrastive Learning}

RT-PCL constrains feature representations around low-confidence and confidence-discrepant regions. It treats transition bands between reliability roles as query-mask candidates, rather than as ground-truth anatomical boundaries. The foreground gate suppresses obvious background responses, and network-specific class prototypes provide class anchors within each feature space. Let $\bar{\mathbf{p}}(v)=\frac{1}{2}(\mathbf{p}_1(v)+\mathbf{p}_2(v))$ denote the averaged prediction of the two networks.
We first extract reliability-transition bands from $\Omega^{UU}$, $\Omega^{1\leftarrow2}$, and $\Omega^{2\leftarrow1}$. The foreground gate then produces the foreground-gated mask $\Omega^{FG}$, yielding foreground-adjacent query regions:
\begin{equation}
\begin{aligned}
&\partial\Omega
= \operatorname{Dilate}(\Omega,r)\oplus\operatorname{Erode}(\Omega,r),\\
&\Omega^{RT}
= \partial\Omega^{UU}\cup\partial\Omega^{1\leftarrow2}\cup\partial\Omega^{2\leftarrow1},\\
&\Omega^{FG}
= \operatorname{Dilate}\!\left(\{v\mid \arg\max_c\bar{p}^{c}(v)\ne0\},r_f\right),\\
&\Omega^{Q}
= \Omega^{RT}\cap\Omega^{FG}.
\end{aligned}
\label{eq:rtquery}
\end{equation}
Here, $\Omega^{RT}$, $\Omega^{FG}$, and $\Omega^{Q}$ denote the reliability-transition band, foreground-gated mask, and final query mask, respectively.
The query mask only represents spatial locations.

The query label is obtained from the detached ensemble prediction:
\begin{equation}
\hat{y}(v)=\arg\max_c \bar{p}^{c}(v),\quad
\mathcal{Q}_{s}
=\{(\mathbf{z}_{s}^{u}(v),\hat{y}(v))\mid v\in\Omega^{Q}\}.
\label{eq:queryfeature}
\end{equation}
This hard pseudo-label is only used as the class index in the contrastive term.
The foreground gate is used to suppress obvious background responses, while prototype selection reduces unreliable positive anchors.

To avoid ambiguity between unaligned feature spaces, RT-PCL constructs network-specific class prototypes. For the $s$-th network and class $c$, the reliable prototype candidate set is
\begin{equation}
\begin{aligned}
\mathcal{P}_{s,c}
=\{\,\mathbf{z}_{s}^{u}(v)\mid
&\,v\in\Omega^{CC},\ \hat{y}_1(v)=\hat{y}_2(v)=c,\\
&\,\max_k\bar{p}^{k}(v)\ge\tau_p\,\}.
\end{aligned}
\label{eq:protoset}
\end{equation}
Here, $\tau_p$ controls the confidence of prototype sources.
In training, candidate pixels of each class are sampled in a class-balanced manner; if a class has insufficient candidates, it is skipped from prototype contrast in the current batch.
The class prototype is defined as
\begin{equation}
\begin{aligned}
\mu_{s,c}
&= \frac{1}{|\mathcal{P}_{s,c}|}\sum_{\mathbf{z}\in\mathcal{P}_{s,c}}\mathbf{z},\\
\mu_{s,c}
&\leftarrow \frac{\mu_{s,c}}{\|\mu_{s,c}\|_2}.
\end{aligned}
\label{eq:prototype}
\end{equation}
For query features from the $s$-th network, RT-PCL pulls each query toward the corresponding prototype in the same feature space and separates it from other class prototypes:
\begin{equation}
\begin{aligned}
\mathcal{L}_{RT}^{(s)}
&= -\mathbb{E}_{v\in\Omega^{Q}}\log
\frac{\exp\!\left(\simf(\mathbf{z}_{s}^{u}(v),\mu_{s,\hat{y}(v)})/T\right)}
{\sum_{c\in\mathcal{C}_{s}}
\exp\!\left(\simf(\mathbf{z}_{s}^{u}(v),\mu_{s,c})/T\right)},\\
\mathcal{L}_{RT}
&= \frac{1}{2}\left(\mathcal{L}_{RT}^{(1)}+\mathcal{L}_{RT}^{(2)}\right).
\end{aligned}
\label{eq:frtpcl}
\end{equation}
Here, $T$ denotes the temperature parameter, and $\mathcal{C}_{s}$ denotes the set of prototype classes available for the $s$-th network in the current batch.
By constructing prototypes within each network, RT-PCL avoids directly sharing class prototypes across unaligned feature spaces and converts complementary dual-network responses into discriminative feature constraints.

\subsection{High-confidence Context Refinement}

HCR exploits stable supervisory information in mutually confident regions to complement RT-PCL, which mainly focuses on reliability-transition regions. Let $\hat{y}_s(v)=\arg\max_c p_s^c(v)$ denote the predicted class of the $s$-th network, and let $\hat{y}(v)=\arg\max_c\bar{p}^c(v)$ denote the ensemble pseudo-label. The pseudo-foreground set and its local boundary band are
\begin{equation}
\begin{aligned}
\Omega^{PF}
&= \{v\mid \hat{y}(v)\ne0\},\\
\partial\hat{Y}
&= \operatorname{Dilate}(\Omega^{PF},r_i)\oplus
\operatorname{Erode}(\Omega^{PF},r_i).
\end{aligned}
\label{eq:pseudo_boundary}
\end{equation}
We then select reliable interior pixels that are prediction-consistent, high-confidence, foreground, and away from this boundary band:
\begin{equation}
\begin{aligned}
\Omega^{RI}
&= \{v\in\Omega^{CC}\mid
\hat{y}_1(v)=\hat{y}_2(v)=\hat{y}(v),\\
&\qquad \hat{y}(v)\ne0,\ 
\max_c\bar{p}^{c}(v)\ge\tau_i,\ 
v\notin\partial\hat{Y}\}.
\end{aligned}
\label{eq:ri}
\end{equation}
For this region, we impose classification and overlap constraints using the ensemble pseudo-label:
\begin{equation}
\mathcal{L}_{ri}=\sum_{s=1}^{2}\left[
\mathcal{L}_{ce}\!\left(f_s(x^u),\hat{y};\Omega^{RI}\right)+
\mathcal{L}_{dice}\!\left(\mathbf{p}_s,\hat{y};\Omega^{RI}\right)
\right].
\label{eq:ri_loss}
\end{equation}

HCR further selects trusted dense blocks according to a local reliability score:
\begin{equation}
\mathcal{B}^{T}=\{B_k\mid R(B_k)\ge\tau_b\}.
\label{eq:dense}
\end{equation}
Here, $R(B_k)$ is determined by the mutually confident ratio, average confidence, and foreground occupancy within the local block.
For selected trusted blocks, we impose local dense classification using the ensemble pseudo-label:
\begin{equation}
\mathcal{L}_{tdb}
=\sum_{s=1}^{2}\mathcal{L}_{ce}\!\left(f_s(x^u),\hat{y};\mathcal{B}^{T}\right).
\label{eq:tdb_loss}
\end{equation}

The final training objective is:
\begin{equation}
\mathcal{L}
= \sum_{s=1}^{2}\mathcal{L}_{sup}^{(s)}
+\mathcal{L}_{RR}
+\lambda_{P}(t)\mathcal{L}_{RT}
+\lambda_{H}\mathcal{L}_{HCR}.
\label{eq:total_loss}
\end{equation}
Here, $\mathcal{L}_{sup}^{(s)}$, $\mathcal{L}_{RT}$, and $\mathcal{L}_{HCR}$ denote the supervised segmentation loss, RT-PCL prototype contrastive loss, and HCR constraints, respectively.

\section{Experiments and Results}

\subsection{Datasets}

The ACDC dataset contains short-axis cardiac cine-MRI scans from 100 patients~\cite{bernard2018deep}.
Following the common semi-supervised setting, 70 cases are used as the training pool, 10 cases are used for validation, and 20 cases are used for testing.
The segmentation targets include the right ventricle, myocardium, and left ventricle.

The MVAA-CT dataset is derived from Task 1 of the MVAA 2026 challenge~\cite{mvaa2026challenge}, containing 27 labeled mitral valve CT volumes and 1040 official unlabeled CT volumes.
We split the 27 labeled cases into 20 training-pool cases and 7 validation cases.
For low-label experiments, we fix the 7 validation cases and construct the 2-labeled and 4-labeled settings within the 20 training-pool cases.
In each setting, labeled cases are randomly sampled five times, while the remaining training-pool cases are used as unlabeled data by hiding their annotations, together with the 1040 official unlabeled CT volumes.
All main comparison results on MVAA-CT are reported as the mean and standard deviation over the five labeled-case samplings.
Compared with relatively compact chamber structures, mitral valve targets are more foreground-sparse and thinner, making local continuity and surface boundaries more sensitive for subsequent structure quantification.

\subsection{Implementation Details}

U-Net~\cite{ronneberger2015unet} and 3D U-Net~\cite{cicek2016threedunet} serve as the backbones for ACDC and MVAA-CT, respectively.
ACDC uses $256\times256$ inputs with a batch size of 16 (8 labeled and 8 unlabeled images), whereas MVAA-CT uses $48\times128\times128$ patches with a batch size of 4 (2 labeled and 2 unlabeled patches).
Both peer networks are optimized with AdamW at an initial learning rate of $1\times10^{-4}$, for 30K iterations on ACDC and 15K on MVAA-CT.
RT-PCL is activated after 3K iterations to avoid unstable early prototypes.
We evaluate DSC (Dice similarity coefficient), Jaccard, 95HD (95\% Hausdorff distance), and ASD (average surface distance).
Reproduced methods follow their public implementations or original descriptions with necessary dataset-specific adaptations, using identical data splits, preprocessing, and evaluation metrics.
All semi-supervised methods on MVAA-CT use the same unlabeled pool, including hidden-label training-pool volumes and 1040 official unlabeled CT volumes.

\begin{table}[!t]
\centering
\caption{Comparison on the ACDC test set. Best and second-best results are bold and underlined; $\dagger$ denotes reproduced results.}
\label{tab:acdc_main}
\scriptsize
\setlength{\tabcolsep}{0.35pt}
\renewcommand{\arraystretch}{0.90}
\begin{tabular*}{\columnwidth}{@{\extracolsep{\fill}}l|cc@{\hspace{2pt}}|cccc@{}}
\toprule
Method & Labeled & Unlabeled & DSC$\uparrow$ & Jaccard$\uparrow$ & 95HD$\downarrow$ & ASD$\downarrow$ \\
\midrule
\multirow{3}{*}{U-Net\reimpl~\cite{ronneberger2015unet}} & 7(10\%)   & 0       & 77.07 & 65.14 & 8.47 & 2.65 \\
                                      & 14(20\%)  & 0       & 84.12 & 74.21 & 6.46 & 2.03 \\
                                      & 70(All)   & 0       & 91.47 & 84.68 & 1.40 & 0.34 \\
\midrule
CPS\reimpl~\cite{chen2021cps}      & \multirow{9}{*}{7(10\%)}  & \multirow{9}{*}{63(90\%)} & 86.74 & 77.43 & 6.64 & 1.80 \\
MC-Net+\reimpl~\cite{wu2022mcnet} &                           &                           & 86.31 & 76.71 & 8.23 & 2.20 \\
BCP\reimpl~\cite{bai2023bcp}      &                           &                           & 88.23 & 79.71 & 4.54 & 1.41 \\
DiffRect\reimpl~\cite{liu2024diffrect} &                         &                           & \second{89.53} & \second{81.57} & 4.33 & 1.21 \\
MLRP\reimpl~\cite{su2024mlrp}    &                           &                           & 87.99 & 79.28 & 3.28 & 0.90 \\
AdaMix\reimpl~\cite{shen2026adamix} &                         &                           & 89.08 & 80.90 & 1.82 & 0.71 \\
M3HL\reimpl~\cite{liu2025m3hl}    &                           &                           & 89.35 & 81.35 & \second{1.54} & \second{0.61} \\
\addlinespace[0.45pt]
W2SPC~\cite{yang2025w2spc}   &                           &                           & 88.92 & 79.65 & 4.32 & 1.19 \\
\addlinespace[0.45pt]
\textbf{R2CL(Ours)}       &                           &                           & \best{89.93} & \best{82.16} & \best{1.43} & \best{0.47} \\
\midrule
CPS\reimpl~\cite{chen2021cps}      & \multirow{9}{*}{14(20\%)} & \multirow{9}{*}{56(80\%)} & 88.48 & 80.28 & 5.47 & 1.59 \\
MC-Net+\reimpl~\cite{wu2022mcnet} &                           &                           & 87.94 & 79.50 & 5.77 & 1.66 \\
BCP\reimpl~\cite{bai2023bcp}      &                           &                           & 89.20 & 81.13 & 4.80 & 1.56 \\
DiffRect\reimpl~\cite{liu2024diffrect} &                         &                           & 89.92 & 82.27 & 3.44 & 0.93 \\
MLRP\reimpl~\cite{su2024mlrp}    &                           &                           & 88.76 & 80.64 & 4.49 & 1.16 \\
AdaMix\reimpl~\cite{shen2026adamix} &                         &                           & 89.76 & 81.27 & \second{1.71} & 0.63 \\
M3HL\reimpl~\cite{liu2025m3hl}    &                           &                           & 89.99 & \second{82.38} & 2.04 & \second{0.55} \\
\addlinespace[0.45pt]
W2SPC~\cite{yang2025w2spc}   &                           &                           & \second{90.23} & 81.24 & 4.01 & 1.06 \\
\addlinespace[0.45pt]
\textbf{R2CL(Ours)}       &                           &                           & \best{90.75} & \best{83.53} & \best{1.37} & \best{0.42} \\
\bottomrule
\end{tabular*}
\end{table}

\begin{table}[!b]
\centering
\caption{Five repeated labeled-case sampling results on the MVAA-CT validation set. Results are reported as mean $\pm$ standard deviation. Best and second-best mean values are bold and underlined; $\dagger$ denotes reproduced results.}
\label{tab:mvaa_main}
\scriptsize
\setlength{\tabcolsep}{0.18pt}
\renewcommand{\arraystretch}{0.90}
\begin{tabular*}{\columnwidth}{@{\extracolsep{\fill}}l|cc@{\hspace{2pt}}|cccc@{}}
\toprule
Method & Labeled & Unlabeled & DSC$\uparrow$ & Jaccard$\uparrow$ & 95HD$\downarrow$ & ASD$\downarrow$ \\
\midrule
\multirow{2}{*}{3D U-Net\reimpl~\cite{cicek2016threedunet}} & 2   & 0       & 74.55$\pm$3.83 & 60.13$\pm$4.90 & 4.56$\pm$2.10 & 1.18$\pm$0.44 \\
                           & 4   & 0       & 81.06$\pm$0.93 & 68.34$\pm$1.27 & 1.83$\pm$0.43 & 0.59$\pm$0.14 \\
\midrule
MC-Net+\reimpl~\cite{wu2022mcnet} & \multirow{9}{*}{2} & \multirow{9}{*}{1058} & 73.28$\pm$2.21 & 58.92$\pm$3.36 & 3.23$\pm$1.02 & 0.74$\pm$0.25 \\
CAML\reimpl~\cite{gao2023caml}   &                          &                           & 72.33$\pm$3.17 & 57.25$\pm$4.02 & 3.79$\pm$1.47 & 0.80$\pm$0.42 \\
BCP\reimpl~\cite{bai2023bcp}      &                          &                           & \second{75.92$\pm$1.98} & \second{61.22$\pm$2.56} & \second{2.58$\pm$0.67} & \second{0.73$\pm$0.16} \\
MLRP\reimpl~\cite{su2024mlrp}    &                          &                           & 71.46$\pm$3.52 & 56.94$\pm$4.13 & 4.26$\pm$1.67 & 1.15$\pm$0.94 \\
M3HL\reimpl~\cite{liu2025m3hl}    &                          &                           & 71.70$\pm$3.55 & 56.35$\pm$4.39 & 4.22$\pm$1.35 & 1.19$\pm$0.27 \\
PICK\reimpl~\cite{zeng2025pick}     &                          &                           & 73.55$\pm$5.21 & 59.16$\pm$5.53 & 3.42$\pm$1.24 & 0.77$\pm$0.18 \\
LeFeD\reimpl~\cite{zeng2025lefed}     &                          &                           & 75.48$\pm$4.33 & 61.14$\pm$5.35 & 3.23$\pm$1.31 & 0.94$\pm$0.33 \\
AdaMix\reimpl~\cite{shen2026adamix} &                         &                           & 72.59$\pm$3.02 & 57.26$\pm$3.67 & 4.11$\pm$2.48 & 1.52$\pm$0.49 \\
\addlinespace[0.45pt]
\textbf{R2CL(Ours)}       &                          &                           & \best{78.48$\pm$1.69} & \best{64.89$\pm$2.23} & \best{2.08$\pm$0.45} & \best{0.62$\pm$0.08} \\
\midrule
MC-Net+\reimpl~\cite{wu2022mcnet} & \multirow{9}{*}{4} & \multirow{9}{*}{1056} & 80.32$\pm$1.44 & 66.75$\pm$1.56 & 1.98$\pm$0.52 & 0.65$\pm$0.28 \\
CAML\reimpl~\cite{gao2023caml}   &                          &                           & 79.87$\pm$1.22 & 66.43$\pm$1.37 & 2.02$\pm$0.58 & 0.72$\pm$0.26 \\
BCP\reimpl~\cite{bai2023bcp}      &                          &                           & 79.43$\pm$1.07 & 66.16$\pm$1.39 & 2.03$\pm$0.41 & 0.64$\pm$0.17 \\
MLRP\reimpl~\cite{su2024mlrp}    &                          &                           & 80.99$\pm$1.78 & 68.28$\pm$2.44 & 1.66$\pm$0.56 & 0.53$\pm$0.22 \\
M3HL\reimpl~\cite{liu2025m3hl}    &                          &                           & 78.02$\pm$0.51 & 65.56$\pm$1.61 & 3.35$\pm$0.69 & 0.96$\pm$0.26 \\
PICK\reimpl~\cite{zeng2025pick}     &                          &                           & 80.08$\pm$1.13 & 67.10$\pm$1.36 & 2.04$\pm$0.62 & 0.65$\pm$0.20 \\
LeFeD\reimpl~\cite{zeng2025lefed}     &                          &                           & \second{81.27$\pm$1.19} & \second{68.73$\pm$1.59} & 1.62$\pm$0.47 & \second{0.51$\pm$0.19} \\
AdaMix\reimpl~\cite{shen2026adamix} &                         &                           & 81.04$\pm$1.60 & 68.71$\pm$2.26 & \second{1.44$\pm$0.43} & 0.55$\pm$0.23 \\
\addlinespace[0.45pt]
\textbf{R2CL(Ours)}       &                          &                           & \best{82.66$\pm$0.80} & \best{70.63$\pm$1.12} & \best{1.29$\pm$0.37} & \best{0.41$\pm$0.15} \\
\bottomrule
\end{tabular*}
\end{table}

\begin{figure*}[!t]
\centering
\includegraphics[width=0.95\textwidth]{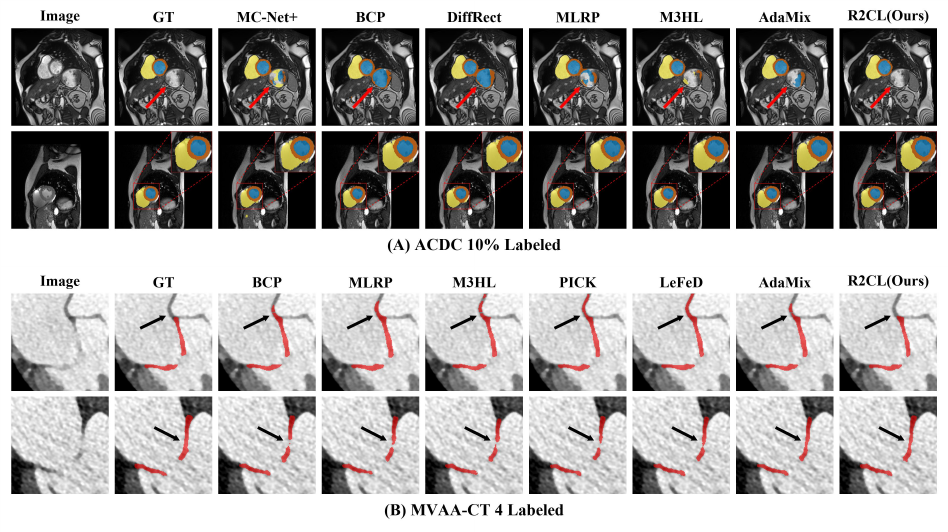}
\caption{Segmentation visualizations on ACDC test cases and MVAA-CT validation cases. Panel (A) shows ACDC with 7 labeled cases (10\%), and panel (B) shows MVAA-CT under the 4-labeled setting. Comparisons are intended within each dataset.}
\label{fig:acdc_qualitative}
\end{figure*}

\subsection{Comparison with Existing Methods}

\textbf{1) Results on ACDC:}
Table~\ref{tab:acdc_main} reports the comparison results on ACDC.
Under both 10\% and 20\% labeled settings, R2CL achieves the best results across all four metrics among the compared semi-supervised methods.
With 10\% labels, R2CL achieves 89.93 DSC, outperforming DiffRect~\cite{liu2024diffrect} by 0.40, and reduces 95HD to 1.43, compared with 1.54 for M3HL~\cite{liu2025m3hl}.
With 20\% labels, R2CL achieves 90.75 DSC and reduces 95HD and ASD to 1.37 and 0.42, respectively, indicating that modeling difficult boundary-adjacent regions helps reduce boundary errors.

\textbf{2) Results on MVAA-CT:}
Table~\ref{tab:mvaa_main} reports the average comparison results over five labeled-case samplings on MVAA-CT.
Under both 2- and 4-labeled settings, R2CL achieves the best average results across all four metrics.
Under the 2-labeled setting, R2CL achieves $78.48\pm1.69$ DSC, outperforming BCP~\cite{bai2023bcp} by 2.56 percentage points, while reducing 95HD and ASD to $2.08\pm0.45$ and $0.62\pm0.08$, respectively.
Under the 4-labeled setting, R2CL achieves $82.66\pm0.80$ DSC and $0.41\pm0.15$ ASD, indicating improved segmentation performance with limited annotations.

MVAA-CT presents a more foreground-sparse segmentation problem than ACDC, with foreground voxels occupying approximately 1.92\% of labeled MVAA-CT volumes compared with 3.79\% for ACDC.
The thin and variable mitral valve structures are susceptible to discontinuities and false-positive predictions, particularly under the 2-labeled setting.
By combining role-specific supervision with foreground-adjacent reliability-transition learning and high-confidence context refinement, R2CL may better address these challenges, as reflected in its improvements on MVAA-CT.

\begin{table}[!t]
\centering
\caption{Stepwise component ablation on ACDC and MVAA-CT. ACDC uses the 7-labeled setting, and MVAA-CT uses the 2-labeled setting. Component ablation is conducted on the same default split.}
\label{tab:ablation}
\scriptsize
\setlength{\tabcolsep}{1.2pt}
\renewcommand{\arraystretch}{1.05}
\begin{tabular}{@{}llcccc@{}}
\toprule
Dataset & Variant & DSC$\uparrow$ & Jaccard$\uparrow$ & 95HD$\downarrow$ & ASD$\downarrow$ \\
\midrule
\multirow{5}{*}{ACDC}
& Baseline & 86.54 & 77.32 & 5.96 & 1.79 \\
& +RR-MCL & 88.39 & 79.78 & 3.23 & 0.68 \\
& +RR-MCL+RT-PCL & 89.22 & 81.32 & 1.85 & 0.64 \\
& +RR-MCL+HCR & 89.56 & 81.61 & 2.41 & 0.66 \\
& \textbf{R2CL(Ours)} & \best{89.93} & \best{82.16} & \best{1.43} & \best{0.47} \\
\midrule
\multirow{5}{*}{MVAA-CT}
& Baseline & 71.93 & 56.45 & 4.98 & 1.28 \\
& +RR-MCL & 74.84 & 60.25 & 2.59 & 0.82 \\
& +RR-MCL+RT-PCL & 76.06 & 61.71 & 1.93 & 0.62 \\
& +RR-MCL+HCR & 77.23 & 63.26 & 2.28 & 0.67 \\
& \textbf{R2CL(Ours)} & \best{78.10} & \best{64.35} & \best{1.71} & \best{0.55} \\
\bottomrule
\end{tabular}
\end{table}

\textbf{3) Visual comparison:}
Fig.~\ref{fig:acdc_qualitative} presents segmentation visualizations on ACDC and MVAA-CT under the corresponding low-label settings.
The figure uses 7 labeled cases (10\%) for ACDC and the 4-labeled setting for MVAA-CT.
Visual comparisons are made only within each dataset.
In the ACDC examples, several compared methods produce small false-positive regions in the background adjacent to the true foreground, whereas R2CL better suppresses such responses.
On MVAA-CT, R2CL shows more stable performance in false-positive suppression and elongated-structure continuity.

\subsection{Ablation Analysis}

\textbf{1) Component ablation:}

Table~\ref{tab:ablation} evaluates RR-MCL, RT-PCL, and HCR against a dual-network cross-pseudo-supervision baseline on the fixed ACDC 7-labeled and MVAA-CT 2-labeled splits.
Since RT-PCL and HCR rely on the reliability regions generated by RR-MCL, they are evaluated after introducing RR-MCL.
On ACDC, RR-MCL improves DSC from 86.54 to 88.39 and reduces 95HD from 5.96 to 3.23.
Adding RT-PCL further reduces 95HD to 1.85, whereas adding HCR to RR-MCL yields larger improvements in region-overlap metrics.
The full R2CL achieves 89.93 DSC and 1.43 95HD.

On MVAA-CT, the components also provide improvements with different emphases.
RR-MCL increases DSC from 71.93 to 74.84; adding RT-PCL reduces 95HD from 2.59 to 1.93, while adding HCR increases DSC to 77.23.
These results indicate that low-confidence regions, reliability-transition regions, and mutually confident regions provide complementary learning signals.

\begin{figure}[!tbp]
\centering
\includegraphics[width=\columnwidth]{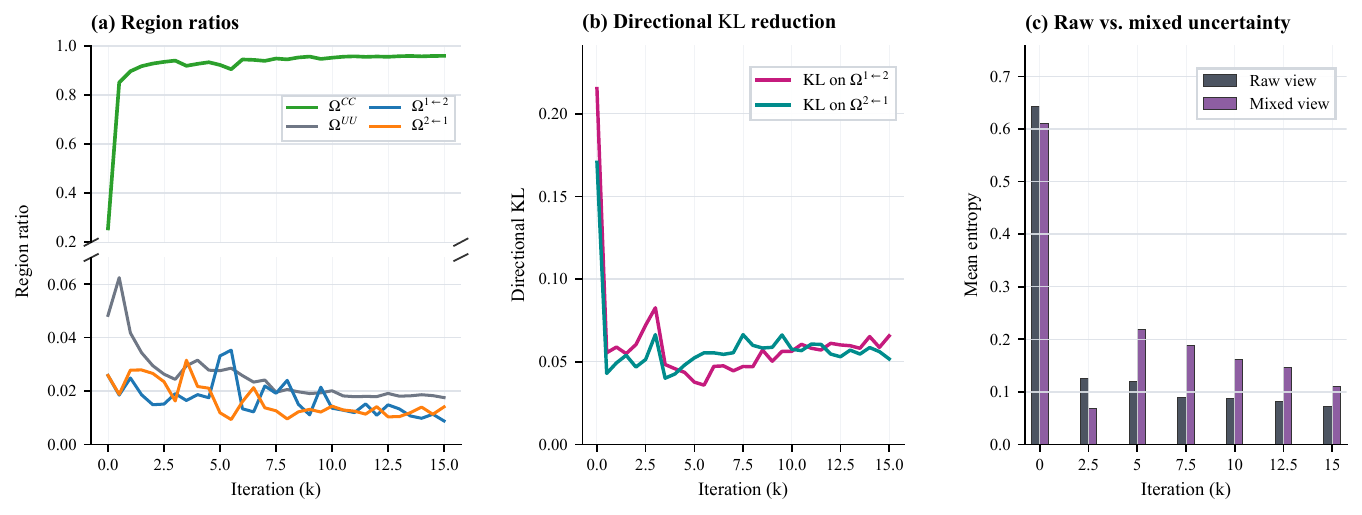}
\caption{RR-MCL training dynamics on MVAA-CT: region proportions, directional KL reduction, and raw/mixed-view uncertainty. Here, $\Omega^{CC}$, $\Omega^{UU}$, $\Omega^{1\leftarrow2}$, and $\Omega^{2\leftarrow1}$ denote mutually confident, mutually uncertain, network-2-guided, and network-1-guided regions, respectively.}
\label{fig:frmcl_region_role_mvaa}
\end{figure}

\textbf{2) RR-MCL training dynamics analysis:}
Fig.~\ref{fig:frmcl_region_role_mvaa} analyzes how RR-MCL converts reliability regions into training roles on MVAA-CT.
Fig.~\ref{fig:frmcl_region_role_mvaa}(a) shows that $\Omega^{CC}$ dominates after stabilization, while sparse but persistent difficult regions remain.
Fig.~\ref{fig:frmcl_region_role_mvaa}(b) shows decreasing directional KL, and Fig.~\ref{fig:frmcl_region_role_mvaa}(c) shows that the mixed view maintains higher uncertainty after stabilization.

\begin{figure}[!tbp]
\centering
\includegraphics[width=0.92\columnwidth]{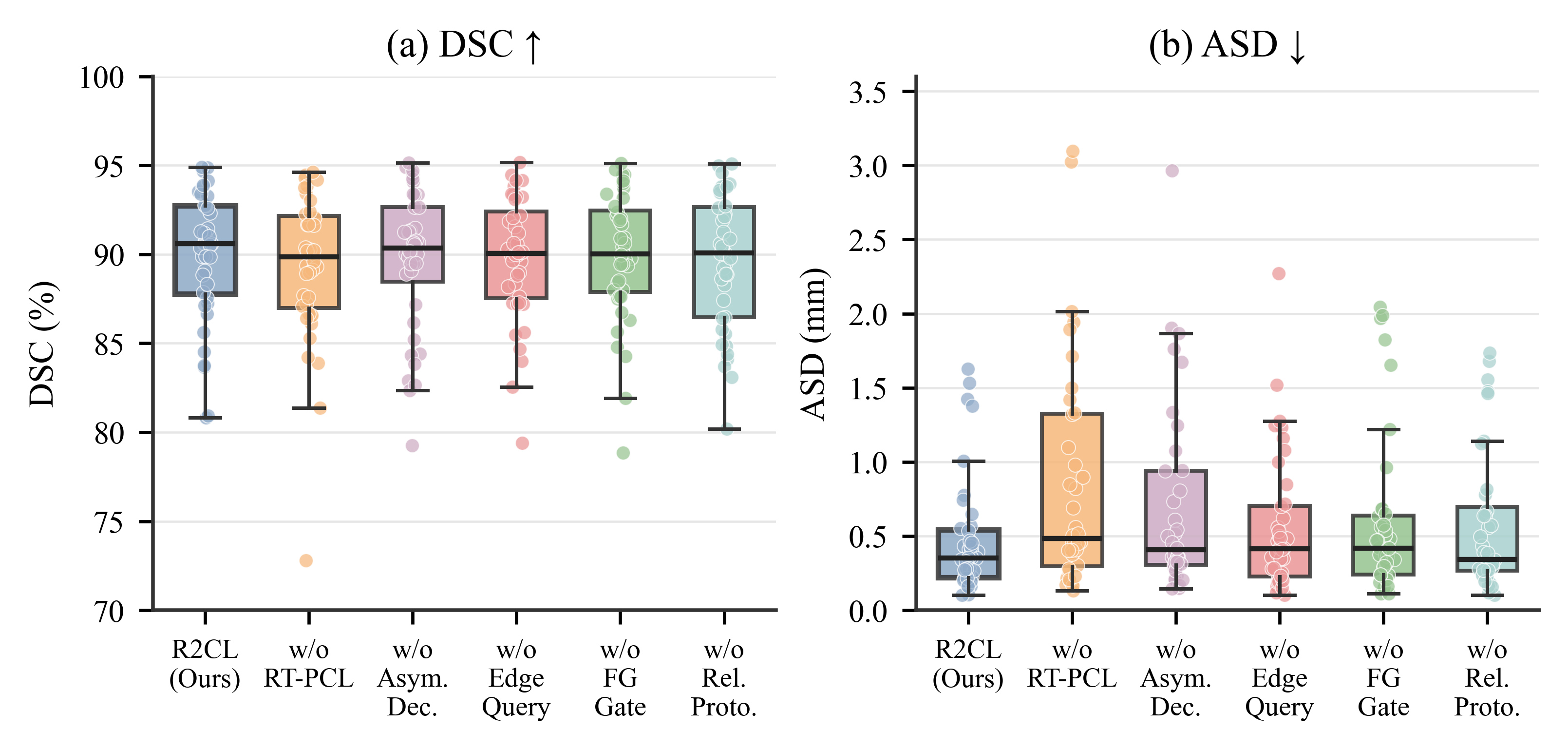}
\caption{Per-case DSC and ASD distributions of the full model and RT-PCL ablation variants on the ACDC test set.}
\label{fig:per_case_boxplot}
\end{figure}

\needspace{4\baselineskip}
\textbf{3) Case-level stability:}
Fig.~\ref{fig:per_case_boxplot} compares per-sample DSC and ASD distributions across six variants on 40 test volumes/frames from 20 ACDC patients.
The variants include R2CL(Ours), w/o RT-PCL, w/o Asym. Dec., w/o RT Query, w/o FG Gate, and w/o Rel. Proto.
R2CL maintains a high median DSC and a lower, more compact ASD distribution, whereas removing RT-PCL or its key designs increases boundary errors in some cases.

\FloatBarrier
\section{Conclusion}

We presented R2CL, a reliability-role collaborative learning framework for semi-supervised cardiac image segmentation.
Rather than applying uniform supervision to all unlabeled pixels, R2CL models unlabeled regions according to their reliability roles and exploits them through RR-MCL, RT-PCL, and HCR.
Experiments on ACDC and MVAA-CT showed consistent improvements in overlap and boundary-related metrics, with larger gains on the foreground-sparse MVAA-CT dataset.
These results indicate that reliability-role modeling can improve segmentation robustness under limited annotations and may benefit downstream clinical applications such as anatomical quantification and morphology assessment.
Future work will explore adaptive reliability estimation and broader medical segmentation scenarios.

\begingroup
\balance
\bibliographystyle{IEEEtran}
\bibliography{references}

\begin{thebibliography}{10}
\providecommand{\url}[1]{#1}
\csname url@samestyle\endcsname
\providecommand{\newblock}{\relax}
\providecommand{\bibinfo}[2]{#2}
\providecommand{\BIBentrySTDinterwordspacing}{\spaceskip=0pt\relax}
\providecommand{\BIBentryALTinterwordstretchfactor}{4}
\providecommand{\BIBentryALTinterwordspacing}{\spaceskip=\fontdimen2\font plus
\BIBentryALTinterwordstretchfactor\fontdimen3\font minus
  \fontdimen4\font\relax}
\providecommand{\BIBforeignlanguage}[2]{{%
\expandafter\ifx\csname l@#1\endcsname\relax
\typeout{** WARNING: IEEEtran.bst: No hyphenation pattern has been}%
\typeout{** loaded for the language `#1'. Using the pattern for}%
\typeout{** the default language instead.}%
\else
\language=\csname l@#1\endcsname
\fi
#2}}
\providecommand{\BIBdecl}{\relax}
\BIBdecl

\bibitem{ronneberger2015unet}
O.~Ronneberger, P.~Fischer, and T.~Brox, ``{U-Net}: Convolutional networks for
  biomedical image segmentation,'' in \emph{Proc. MICCAI}, 2015, pp. 234--241.

\bibitem{isensee2021nnunet}
F.~Isensee, P.~F. Jaeger, S.~A.~A. Kohl, J.~Petersen, and K.~H. Maier-Hein,
  ``{nnU-Net}: A self-configuring method for deep learning-based biomedical
  image segmentation,'' \emph{Nature Methods}, vol.~18, pp. 203--211, 2021.

\bibitem{chen2024transunet}
J.~Chen, Y.~Lu, Q.~Yu, X.~Luo, E.~Adeli, Y.~Wang, L.~Lu, A.~L. Yuille, and
  Y.~Zhou, ``{TransUNet}: Rethinking the {U-Net} architecture design for
  medical image segmentation through the lens of transformers,'' \emph{Medical
  Image Analysis}, vol.~97, p. 103280, 2024.

\bibitem{hatamizadeh2022swinunetr}
A.~Hatamizadeh, V.~Nath, Y.~Tang, D.~Yang, H.~R. Roth, and D.~Xu, ``{Swin
  UNETR}: {Swin} transformers for semantic segmentation of brain tumors in
  {MRI} images,'' in \emph{Brainlesion: Glioma, Multiple Sclerosis, Stroke and
  Traumatic Brain Injuries}, ser. Lecture Notes in Computer Science, vol.
  12962, 2022, pp. 272--284.

\bibitem{lee2013pseudo}
D.-H. Lee, ``Pseudo-label: The simple and efficient semi-supervised learning
  method for deep neural networks,'' in \emph{ICML Workshop on Challenges in
  Representation Learning}, 2013.

\bibitem{tarvainen2017mean}
A.~Tarvainen and H.~Valpola, ``Mean teachers are better role models:
  Weight-averaged consistency targets improve semi-supervised deep learning
  results,'' in \emph{Proc. NeurIPS}, 2017, pp. 1195--1204.

\bibitem{chen2021cps}
X.~Chen, Y.~Yuan, G.~Zeng, and J.~Wang, ``Semi-supervised semantic segmentation
  with cross pseudo supervision,'' in \emph{Proc. CVPR}, 2021, pp. 2613--2622.

\bibitem{miyato2019vat}
T.~Miyato, S.~ichi Maeda, M.~Koyama, and S.~Ishii, ``Virtual adversarial
  training: A regularization method for supervised and semi-supervised
  learning,'' \emph{IEEE Trans. Pattern Anal. Mach. Intell.}, vol.~41, no.~8,
  pp. 1979--1993, 2019.

\bibitem{berthelot2019mixmatch}
D.~Berthelot, N.~Carlini, I.~Goodfellow, N.~Papernot, A.~Oliver, and C.~Raffel,
  ``{MixMatch}: A holistic approach to semi-supervised learning,'' in
  \emph{Proc. NeurIPS}, 2019, pp. 5049--5059.

\bibitem{wu2022mcnet}
Y.~Wu, Z.~Ge, D.~Zhang, M.~Xu, L.~Zhang, Y.~Xia, and J.~Cai, ``Mutual
  consistency learning for semi-supervised medical image segmentation,''
  \emph{Med. Image Anal.}, vol.~81, p. 102530, 2022.

\bibitem{yu2019uamt}
L.~Yu, S.~Wang, X.~Li, C.-W. Fu, and P.-A. Heng, ``Uncertainty-aware
  self-ensembling model for semi-supervised {3D} left atrium segmentation,'' in
  \emph{Proc. MICCAI}, 2019, pp. 605--613.

\bibitem{bai2023bcp}
Y.~Bai, D.~Chen, Q.~Li, W.~Shen, and Y.~Wang, ``Bidirectional copy-paste for
  semi-supervised medical image segmentation,'' in \emph{Proc. CVPR}, 2023, pp.
  11\,514--11\,524.

\bibitem{lei2025alhvr}
T.~Lei, Z.~Yang, X.~Wang, Y.~Wang, X.~Wang, F.~Sun, and A.~K. Nandi, ``Adaptive
  learning of high-value regions for semi-supervised medical image
  segmentation,'' in \emph{Proc. ICCV}, 2025, pp. 21\,450--21\,459.

\bibitem{luo2018multiview}
G.~Luo, S.~Dong, K.~Wang, W.~Zuo, S.~Cao, and H.~Zhang, ``Multi-views fusion
  {CNN} for left ventricular volumes estimation on cardiac {MR} images,''
  \emph{IEEE Transactions on Biomedical Engineering}, vol.~65, no.~9, pp.
  1924--1934, 2018.

\bibitem{luo2020dcn}
G.~Luo, W.~Wang, C.~Tam, K.~Wang, S.~Cao, H.~Zhang, B.~Chen, and S.~Li,
  ``Dynamically constructed network with error correction for accurate
  ventricle volume estimation,'' \emph{Medical Image Analysis}, vol.~64, p.
  101723, 2020.

\bibitem{luo2020ccn}
G.~Luo, S.~Dong, W.~Wang, K.~Wang, S.~Cao, C.~Tam, H.~Zhang, J.~Howey,
  P.~Ohorodnyk, and S.~Li, ``Commensal correlation network between segmentation
  and direct area estimation for bi-ventricle quantification,'' \emph{Medical
  Image Analysis}, vol.~59, p. 101591, 2020.

\bibitem{blum1998cotrain}
A.~Blum and T.~Mitchell, ``Combining labeled and unlabeled data with
  co-training,'' in \emph{Proc. COLT}, 1998, pp. 92--100.

\bibitem{bernard2018deep}
O.~Bernard, A.~Lalande, C.~Zotti, F.~Cervenansky, X.~Yang \emph{et~al.}, ``Deep
  learning techniques for automatic {MRI} cardiac multi-structures segmentation
  and diagnosis: Is the problem solved?'' \emph{IEEE Trans. Med. Imaging},
  vol.~37, no.~11, pp. 2514--2525, 2018.

\bibitem{mvaa2026challenge}
\BIBentryALTinterwordspacing
{MVAA 2026 Challenge Organizers}, ``{MVAA 2026}: Mitral valve anatomy analysis
  using multimodal imaging data,'' Codabench online challenge benchmark, 2026,
  accessed: Jun. 28, 2026. [Online]. Available:
  \url{https://www.codabench.org/competitions/15662/}
\BIBentrySTDinterwordspacing

\bibitem{cicek2016threedunet}
{\"O}.~{\c{C}}i{\c{c}}ek, A.~Abdulkadir, S.~S. Lienkamp, T.~Brox, and
  O.~Ronneberger, ``{3D U-Net}: Learning dense volumetric segmentation from
  sparse annotation,'' in \emph{Proc. MICCAI}, 2016, pp. 424--432.

\bibitem{liu2024diffrect}
X.~Liu, W.~Li, and Y.~Yuan, ``{DiffRect}: Latent diffusion label rectification
  for semi-supervised medical image segmentation,'' in \emph{Proc. MICCAI},
  2024, pp. 56--66.

\bibitem{su2024mlrp}
J.~Su, Z.~Luo, S.~Lian, D.~Lin, and S.~Li, ``Mutual learning with reliable
  pseudo label for semi-supervised medical image segmentation,'' \emph{Med.
  Image Anal.}, vol.~94, p. 103111, 2024.

\bibitem{shen2026adamix}
Z.~Shen, P.~Cao, J.~Su, J.~Yang, and O.~R. Zaiane, ``Adaptive mix for
  semi-supervised medical image segmentation,'' \emph{Medical Image Analysis},
  vol. 108, p. 103857, 2026.

\bibitem{liu2025m3hl}
Y.~Liu, Z.~Zhang, J.~Yue, W.~Guo, and D.~Li, ``{M3HL}: Mutual mask mix with
  high-low level feature consistency for semi-supervised medical image
  segmentation,'' in \emph{Proc. MICCAI}, 2025, pp. 314--323.

\bibitem{yang2025w2spc}
Y.~Yang, G.~Sun, T.~Zhang, R.~Wang, and J.~Su, ``Semi-supervised medical image
  segmentation via weak-to-strong perturbation consistency and edge-aware
  contrastive representation,'' \emph{Med. Image Anal.}, vol. 101, p. 103450,
  2025.

\bibitem{gao2023caml}
S.~Gao, Z.~Zhang, J.~Ma, Z.~Li, and S.~Zhang, ``Correlation-aware mutual
  learning for semi-supervised medical image segmentation,'' in \emph{Proc.
  MICCAI}, 2023, pp. 98--108.

\bibitem{zeng2025pick}
Q.~Zeng, Z.~Lu, Y.~Xie, and Y.~Xia, ``{PICK}: Predict and mask for
  semi-supervised medical image segmentation,'' \emph{Int. J. Comput. Vis.},
  vol. 133, pp. 3296--3311, 2025.

\bibitem{zeng2025lefed}
Q.~Zeng, Y.~Xie, Z.~Lu, M.~Lu, J.~Zhang, and Y.~Xia, ``Consistency-guided
  differential decoding for enhancing semi-supervised medical image
  segmentation,'' \emph{IEEE Trans. Med. Imaging}, vol.~44, no.~1, pp. 44--56,
  2025.

\end{thebibliography}
\endgroup

\end{document}